# A Multi-Agents Architecture to Learn Vision Operators and their Parameters

Issam Qaffou, Mohammed Sadgal, Abdelaziz Elfazziki

Département Informatique,
Faculté des Sciences Semlalia, Université Cadi Ayyad,
Bd Prince My Abdellah BP 2390,
Marrakech, Morocco
{i.qaffou, sadgal, elfazziki}@ucam.ac.ma

**Abstract**

In a vision system, every task needs that the operators to apply should be « well chosen » and their parameters should be also « well adjusted ». The diversity of operators and the multitude of their parameters constitute a big challenge for users. As it is very difficult to make the « right » choice, lack of a specific rule, many disadvantages appear and affect the computation time and especially the quality of results. In this paper we present a multi-agent architecture to learn the best operators to apply and their best parameters for a class of images. Our architecture consists of three types of agents: User Agent, Operator Agent and Parameter Agent. The User Agent determines the phases of treatment, a library of operators and the possible values of their parameters. The Operator Agent constructs all possible combinations of operators and the Parameter Agent, the core of the architecture, adjusts the parameters of each combination by treating a large number of images. Through the reinforcement learning mechanism, our architecture does not consider only the system opportunities but also the user preferences.

## 1. Introduction

To accomplish an image processing task (segmentation, detection, object recognition, etc.) the user finds himself faced with a multitude of applicable operators averaging the fixation of values for several parameters. The quality of results depends essentially on the operator chosen and the values assigned to its parameters. The lack of a general rule that guides the user in his choices pushes him usually to use his experience and sometimes his intuition. He, generally, proceeds by trial and error until the identification of a satisfactory result. The problem is already remarkable when the task needs to apply just one operator with several parameters to adjust. However, in the majority of vision tasks, the user is required and sometimes obliged to combine several operators, whose each one has a multitude of parameters to adjust. The user must select operators, adjust their parameters and then test them sequentially on the image. This process is repeated for a long time before deciding on the quality of the result. It's a tedious work with a great waste of time. Often, the user reuses the last combination of operators for his application. But, it is possible that there is a better combination that has not been tested by the user. Indeed, the exploration and the exploitation of all possible combinations constitute a source of errors before talking about the time spent in the operation. To help the user to perform vision tasks, several solutions have been proposed as systems and GUI. For example, Pandore [1], a standardized library of image processing operators, consists of a set of executable programs performing directly on image files. The development of an image processing application is done from the successive execution of operators where outputs of one operator can be used as inputs of some others. To execute an operator, Pandore offers the user a set of parameters to adjust.

Ariane is a data flow visual programming environment customized for monitoring libraries of executable operators, such as the image processing operators' library Pandore. Ariane is a prototyping tool which allows users to program applications graphically by simple selection and linking of operators represented as connectable boxes. Ariane is a graphical interface that looks like a graph editor. The user selects operators from the existing list, and then links them to compose processing chains. Outputs of some operators are used as inputs of some others.

In these solutions, the selection of operators and the adjustment of their parameters are done manually by the user. They are semi-automatic solutions. Despite, the user finds always difficulty to choose the appropriate operators and adjust their parameters in order to find the best result.

Some authors searched to automate completely the process by proposing systems and methods to automatically choose operators to apply in a vision task without the user intervention. Draper proposed ADORE in 2000. It is a system of object recognition based on MDP (Markov Decision Process) to choose, from a current situation, the operator to apply [2]. Draper used a library of ten operators to recognize duplexes in arial images. ADORE is based on a method which is robust theoretically, but it can't always ensure good results because it uses a predefined and limited library of operators, that's without talking about the adjustment of their parameters. Other authors proposed methods to automatically adjust parameters of vision operators.

B.NICKOLAY et al. proposed a method to automatically optimize the parameters of a machine vision system for surface inspection by using specific Evolutionary Algorithms (EA) [3]. A few years later, Taylor [4] proposed a reinforcement learning framework which uses connectionist systems as function approximators to handle the problem of determining the optimal parameters for a computer vision application even in the case of a highly dimensional, continuous parameter space. More recently, Farhang et al. [5] introduced a new method for segmentation of the prostate in transrectal ultrasound images, using a reinforcement learning (RL) scheme. He divided the initial image into sub-images and works on each sub-image in order to reach a good result. In [6] and [7], we proposed a reinforcement learning method to adjust automatically the parameters of vision operators. Despite all these researches and their results, they stay limited to a predefined type of images or depend on some particular conditions. Until today, there is no method robust, sure and automatic which provides the user the appropriate operators and their optimal parameters values depending on the vision task and the class of images. Hence, we need systems that allow, generally and for any vision task, to automatically determine the best combination of operators and their optimal parameters values to apply.

In this paper we present a solution for this problem by proposing a multi-agents architecture based on reinforcement learning to select automatically the best operators to apply in a vision task, that's while adjusting their parameters values without the user intervention. In the second section we present an overview on reinforcement learning, multi-agent systems and their applications in computer vision. The third section details the proposed approach. The forth section discusses the experience and its results. The last section concludes the paper.

## 2. Overview

In this section we present the two key concepts which underlie our approach. In the first subsection, we discuss the reinforcement learning concept and its application in image processing. The second subsection concerns multi-agent systems and their use to accomplish vision tasks.

2.1 Reinforcement Learning

According to the definition of S.Sutto and G.Barto [9], reinforcement learning defines a type of interaction between an agent and its environment. From a real situation « s » in the environment, the agent chooses and executes an action « a » which causes a transition to the stat « s' ». It receives in return a reinforcement signal « r », which is a penalty if the action leads to a failure or a reward if the action is beneficial; a zero signal means the inability to assign a penalty or a reward.

The agent uses then this signal to improve its strategy, action sequence, in order to maximize the accumulation of its future rewards. For this purpose, it must balance exploration and exploitation. The exploration is to test new action, which could lead to higher earnings. Whereas the exploitation consists to apply the best strategy previously acquired. The interaction between the agent and its environment is presented by Fig. 1:

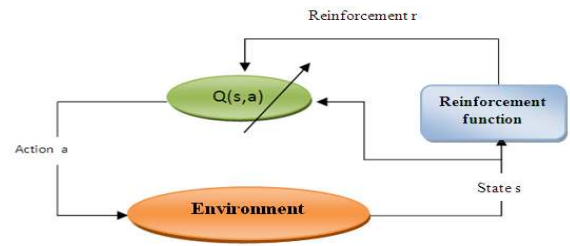

Fig. 1: Reinforcement learning: diagram of interaction agent/environment

- **Influence's factors**

The elements to consider in reinforcement learning are multiple [11] [12]:

**Time:** the time space has different forms, it can be discrete or continuous, finite or infinite, and fixed or random.
Most studies on reinforcement learning use a discrete time space.

**States:** they characterize the situation of an agent and the environment at all times, they can be divided into three forms:
- A relational situation between the agent and the environment (position, etc.);
- A specific situation to the environment (environmental changes);
- An internal situation to the agent (its memory, its captors, etc.).

The three state forms may be present at the same time depending on the processed problem.

**Actions:** the agent chooses an action among the possible actions at each time t; this action may be instantaneous or last until the next time. To each state is associated a set of possible actions.

**Reinforcement signal:** at each time, the interaction produces a reinforcement value r, bounded numeric value, which measures the accuracy of the agent reaction. The goal of the agent is to maximize the "sum" of these reinforcements in time.

Another important element of reinforcement learning is the action policy that defines the agent's behavior at a given time. It maps the visited states to proper actions. There are three common policies, namely Boltzman, ε-greedy, and greedy. The Boltzman policy is a softmax method using a Gibbs distribution that estimates the probability of taking an action a in a given state s. In the greedy policy, the agent always selects greedy actions. In this situation, all actions will not be explored. In the ε-greedy policy the agent selects greedy actions with probability of ε and selects a random action with a probability of 1-ε. The ε-greedy is a popular method to balance exploration and exploitation. Generally, choosing the appropriate policy depends on the application at hand.

Several methods have been proposed based on Markov property: the current state depends only on the previous state. One of the most popular algorithms is Q-learning that we use in our approach choosing the ε-greedy policy.

- **Q-learning**

The Q-learning algorithm is proposed by Watkins [10], reconsidered by S.Sehad [13] as a reinforcement learning method for MDP (Markov Decision Process) when the evolution model is unknown. It's an "off-policy" method. S.Sehad [13] had proposed a model of Q-learning process and underscores the following functions:
- A selection function: from the current situation as perceived by the system, an action is selected and executed based on the knowledge available within the internal memory (that knowledge is stored as a utility value associated with a pair (situation, action)).
- A reinforcement function: after the execution of the action in the real world, the reinforcement function uses the new situation to generate the reinforcement value. This reinforcement is usually a simple value +1, -1 or 0.
- An update function: uses the reinforcement value to adjust the value associated to the state or to the pair (state,action) that has been executed.

The principle of Q-learning is to estimate a function $Q^*$ defined as:
$$Q^*(s,a) = E(r_{t+1} + \gamma V^*(s')) = E(r_{t+1} + \gamma \max_{a'} Q^*(s',a'))$$
Using an asynchronous iterative update given by:
$$Q^\pi_{t+1}(s_t,a_t) = (1-\alpha_t) Q^\pi_t(s,a) + \alpha_t (r_{t+1} + \gamma \max_a Q^\pi_t(s',a'))$$
Where $r_{t+1}$ is the received reinforcement by choosing action "a" in state "s", which makes the process in a new state s' and $\alpha_t$ is the learning rate between 0 and 1; and $\gamma$ is the discount factor: it models the agent preferences according to the reward. In principles, the environment must be explored randomly for a large number of iterations in order that the Q-learning may converge into the optimal Q-function, and only then we can use the optimal policy defined by:
$$\pi^*(s) = \arg\max_{a \in A} Q^*(s,a)$$
Q-learning algorithm:

---
Initialize $Q(s,a)$ arbitrary
Repeat (for each episode)
    Initialize s
    Repeat (for each step of episode)
    Choose $a$ from s using policy derived from Q (e.g., ε-greedy)
    Take action $a$, observe r, s'
$$Q^\pi(s_t,a_t) = (1-\alpha_t) Q^\pi(s,a) + \alpha_t (r + \gamma \max_{a \in A} Q^\pi_t(s',a'))$$
    s←s' ;
    Until s is terminal

---

Some improvements of Q-learning have been proposed. For example minimax-Q [14] and Nash-Q [15]. In Nash-Q, the agent attempts to learn its equilibrium Q-values, starting from an arbitrary guess. Toward this end, the Nash Q-learning agent maintains a model of other agents' Q-values and uses that information to update its own Q-values. The updating rule is based on the expectation that agents would take their equilibrium actions in each state. Minimax-Q called also Friend-or-Foe, has two manners to update the strategies V(s). It can classify the problem as Friend (a global and optimal action exists) or as Foe (there is rather a saddle point). This classification is done by looking at Q-values through the execution of the algorithm. Q-learning is rational and convergent.

2.1.4 Reinforcement learning approaches in image processing

Reinforcement learning techniques have found scope in the field of image processing. For example, a general approach based on reinforcement learning for image segmentation and object recognition is proposed in [16]. This approach adapts parameters of the image segmentation algorithm to the change of the environment. Segmentation parameters are represented by a team of stochastic automata that use connectionist techniques of reinforcement learning. In [17], neural network is trained by reinforcement learning to classify machine parts in a low-resolution image. In another approach, reinforcement learning is introduced for image thresholding where the entropy is used as a reinforcement signal [18]. The application of reinforcement learning in global thresholding is introduced in [19] and [20]. Q-learning algorithm is used to find the optimal threshold for digital images. Two other methods concerning the problem of parameters adjustment are presented in [21] and [22]. They determine membership functions in the contrast adaptation and parameters control for text detection.

2.2 Multi-Agents Systems

- Definitions and principles

The agent concept has been studied for a long time in various disciplines. Multiple definitions of agent have been given depending on the field of application. In our work, we use the definition adopted by Haroun [23] based on M.Wooldridge's works: "an agent is a computer system, situated in some environment, that acts autonomously and flexibly in order to achieve its delegated goals".

A multi-agents system consists of a set of computer processes running simultaneously, so of a set of multiple agents living at the same time, sharing common resources and communicating with each other. The key point of multi-agents systems is the formalization of coordination between agents.

The agents are able to perceive and act on a common environment that they share. Perceptions allow agents to acquire information about their environment evolution, and their actions allow them to change it.

- Multi-agents Approaches in image processing

Many works in image processing use multi-agent systems. In this section we include some of them concerning medical imaging.

In [24], L.Germond proposes different cooperation defined between a deformable model, a multi-agent system and an edge detector. The deformable model is used to extract the brain. From this, several agents are placed on the brain in a way that the agents of the gray matter are placed on the periphery, and those of the white matter are placed inside. The work of N. Richard et al. [25] [26] are intended for the segmentation of brain tissue in white matter, gray matter and cerebrospinal fluid. Agents located on the image work together (cooperation). There are several types of agents: one agent of global control, several agents for local control and agents for segmentation by region-growing, each one of these agents concerns one of the three tissues. In their works [27] [28] Duchesnay et al. perform a multi-agent platform for image segmentation. As a pre-classification, Duchesnay et al. use « split and merge » decomposition, which gives only homogenous regions in the image. An agent is then placed in each homogenous area and the segmentation process based on multi-agents systems begins. They adopt irregular pyramids as an organizational element of the agents' population; the process is made from the basis of the pyramid until its top.

In these works, the segmentation is applied to the scanned breast images.

## 2.3 Learning in Multi-Agents System

In reinforcement learning, an agent is not seen as an individual (like in classical learning) but as an entity that can interact with other entities. In a multi-agents system, learning is centralized if it is globally realized by a single agent which doesn't require any explicit interaction with other environmental agents. In a centralized process, an agent acts as if it is alone.

However a distributed learning within several agents is said decentralized. Agents may have similar or different abilities, and an explicit interaction between agents is necessary to accomplish their goal [29].

## 3. Proposed Approach

Generally, to accomplish a vision task we've to pass through processing phases. Each phase contains a set of operators, usually predefined in a system with their parameters whose some of their values are given by default. Users find themselves faced to a tedious work for choosing the best operator to apply and adjusting its parameters. In our approach we propose a multi-agents architecture which helps automatically the user in his choices (operators and parameters values). The architecture is composed globally by three types of agents. Each one of them is charged to accomplish one task in the process. Fig. 2 shows how these agents are linked.

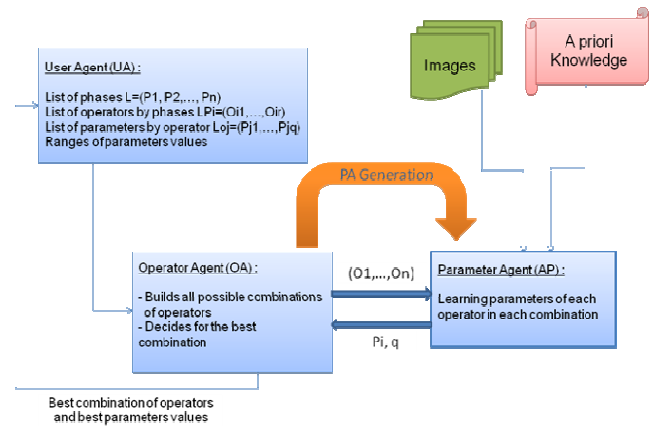

Fig. 2: global schema of the proposed approach. OA proceeds in collaboration with PA.

### 3.1 User Agent (UA)

Depending on the vision task to accomplish, User Agent (UA) gives the list of processing phases. For each phase, it determines a set of possible operators. For each operator it defines parameters to adjust by specifying ranges of their possible values. It also proposes a class of images for learning, on which the system will run, as well as a ground truth for each image. The work of the user agent is necessary so that the operator agent and the parameter agent can proceed.

### 3.2 Operator Agent (OA)

Operator agent proceeds in two steps: the first one is to build, according to the phases determined by UA, all possible combinations of operators. Each combination contains a number of operators equal to the number of phases determined by UA. For each combination, the agent OA generates an agent PA (Parameter Agent) specialized to adjust parameters of its operators. There are then so many agents PA as possible combinations. A global schema for generating agents PA is given in Fig. 3. Each agent PA has its own combination of operators. After adjusting parameters, according to the task at hand, each agent PA returns its combination of operators with the best parameters values. It also returns the result quality of this combination after applying it on a class of images. The second step of the agent OA is to decide among all these operators' combinations which one is the better to apply. The best combination corresponds to this one having the higher result quality. The result is returned to the agent UA.

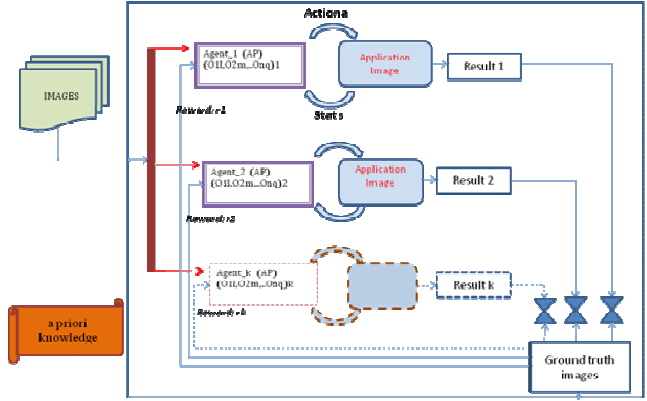

Fig. 3: To each operator's combination is associated an agent PA to adjust their parameters.

Each agent PA uses reinforcement learning to adjust parameters of each operator. It applies actions on a set of images and receives a return which may be a punishment or a reward.
This return is determined depending to a ground truth proposed by an expert (manual processing). More details about how the agents PA proceed are given hereafter.

## 3.3 Parameter Agent (PA)

For each combination of operators, there is an agent PA to adjust parameters of these operators. To do this, a range of values for each parameter and a set of images with their ground truth are given by the agent UA. The agent PA has no prior knowledge about the best parameters values. It proceeds by reinforcement learning to find values giving the best result. Fig. 4 presents a general schema about the functioning of each agent PA. The input image is a processing subject of a series of operators. Each operator has a set of parameters to adjust, and each parameter has a range of possible values.

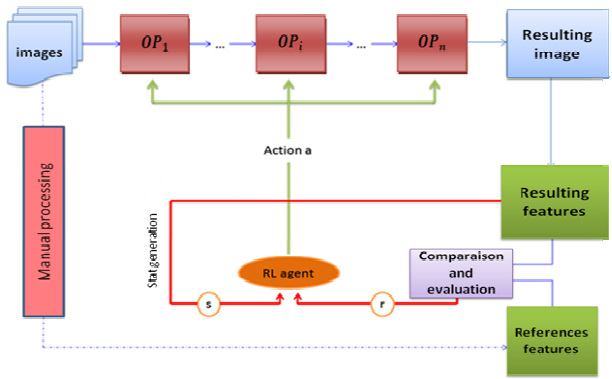

Fig. 4: general schema about the functioning of each agent PA for operator's parameter learning.

The agent PA must find the best parameter values for each operator in order to get the best result. The agent PA uses reinforcement learning as an automatic method to explore all possible values and then exploit the best ones. The agent PA must then define the actions a, states s and reward r. We define actions as all possible combinations of parameters values. States are defined by features describing the image. These features are defined according to the task at hand.

The agent PA chooses an action and applies the operators' combination on the input image and gets a resulting image. For each image there is a ground-truth; it is a resulting image through a manual processing by an expert. The ground-truth represents a reference for the agent PA. To assess the chosen action, the agent PA compares the resulting image with the reference one. It extracts some features from the resulting image and compares them with the same features extracted from the reference image. An evaluation metric is used to assess the result and produce a reward. Each action has its own reward. The best action is the most rewarded. The details about how three components, namely state, action, and reward are defined in our proposed approach are described later (see the next subsections).

### 3.3.1 Defining actions

Generally, all possible combination of parameters values of operators is defined as an action for the agent PA. The set of the actions is then the set of all possible values combination, see fig.4.
Each operator $OP_k$ has a series of parameters: $(P_1^k, P_2^k, ..., P_n^k)$
Each parameter $P_j^k$ has a range of values: $V_j^k = \{V_{j1}^k, V_{j2}^k, ..., V_{jm}^k\}$
An elementary action of the operator $OP_k$ is: $a_k = (u_{j1}^k, ..., u_{jr}^k)$ where $u_{j1}^k \in V_j^k$
An action of the agent PA is defined by the combinations of the elementary actions of operators as it is defined above: $a = (a_1, a_2, ..., a_n)$

### 3.3.2 Defining states

A state is defined by a set of features extracted from the image. From the output of each action, which is an image we extract some features to represent states for the agent PA:

$$s = [\chi_1, \chi_2, ..., \chi_n]$$

Where $\chi_i$ is a feature reflecting the state of the image after the processing.

### 3.3.3 Defining the reward

The return is a reward if the agent PA chooses the right action, else it is a punishment. It is defined according to the quality of the processing result. This quality is assessed by using ground-truth models (manually processed images). To define the return we calculate the similarity between the resulting image and the ground truth image. That is depending to the task at hand. For example, if we use an edge detection approach for image segmentation we would calculate error measures which give global indices about the result quality: over-detection error, under-detection error, localization error, etc. But if we use a region approach we would calculate, for example, errors of Yasnoff [30] or the criterion of Vinet [30], etc. After measuring the similarity's criterions, we assess the result of our system using a weighted sum of the differences of these criterions' scalars:

$$D = \sum_i w_i D_i$$

The weights $w_i$ are chosen according to the importance of each criterion $D_i$.

In our experiments, we've used three error measures: over-detection error, under-detection error and localization error which are formally expressed in the fifth section.

A general form for the reward definition in the proposed approach is presented by:

```
Reward: r= -10, 0 or 10;
if (D < ε) r = +10; f=true;
    else
    if ( (D > ε) && (D < ε + δ) )
        r = 0;
        else r = -10;
    end
end
```

The values 10 and -10 represent respectively the reward and the punishment depending to a predefined threshold. Using the set of images determined by the agent UA, each agent PA returns to the agent OA its combination of operators with the best values of their parameters. It returns also the quality of the result corresponding to the highest reward. The agent OA retrieves then all the combinations it has built with the best parameters values of each operator and the qualities of their results. The agent OA returns to the agent UA the best combination of operators corresponding to the highest quality. Thus it decides for the best combination of operators to apply in order to accomplish à predefined vision task.

In the following section of this paper we discuss the experience and its results for the multi-agent architecture to learn parameters and operators. We tested our approach for segmentation tasks. The operators used in image segmentation differ from an image type to another, they are not necessary the same. Our main goal is to propose a method which determines for each type of images, the best combination of operators to apply to accomplish a predefined task. The example below gives the best combination of operators with the best values of their parameters to use to segment images of traffic signs. This combination is determined from a set of operators proposed by the agent UA.

## 4. Results and Discussion

In the previous sections we described the proposed approach to solve the problem of choosing automatically the best operators and their best parameters values. We presented theoretically our approach as a general method which can be used for any vision task needing operator selection or parameters adjustment or both of them. In this section we test practically the multi agent architecture to choose the right operators and their best parameters values for segmentation tasks. The operators used in image segmentation differ from an image type to another, they are not necessary the same. Our main goal in this section is to show how the proposed approach can determine for a specific type of images, the best combination of operators to apply for their segmentation. The images used in the experience are those of traffic signs.

### 4.1 Segmentation

Segmentation is an important task in the field of computer vision. It is a low-level processing which aims to partitioning an image into homogeneous regions. The goal of segmentation is to extract the entities of an image in order to apply a specific processing and interpret the content of the image. Generally, image segmentation is performed by using one of the two major approaches based on edge extraction (boundaries) or region growing [31] [32].

### 4.1.1 Region-based segmentation

Region-based methods consist in grouping pixels with a common property in a homogeneous zone. The most important methods of this approach are region growing methods [33] [34] [35] and split and merge methods.

### 4.1.2 Boundary-based segmentation

Boundary-based methods are distinguished by measuring intensities local variation representing changes of physical or geometrical properties of an object in the image. Generally, in boundary-based segmentation, we must extract the frontier between regions and then close contours. To extract frontiers, there are two approaches:
- **Gradient approach:** determination of local extrema in the direction of the gradient. Optimal operators were proposed by Canny, Shen and Derriche [35][36][37].
- **Laplacien approach**: determining the zero crossings of the second derivative.

Based on the principles of the two approaches we note that there is a perfect duality between contours and regions.

In the coming sub-sections we'll describe the proposed approach applied for image segmentation. The three used agents will be presented adaptively to the task at hand. Actions, states and reward, the major component of reinforcement learning, will be defined.

### 4.2 The Agent UA

As said above, it defines to the system phases of processing, the operators of each phase and values of their parameters. Below we give examples of the agents UA, OA and PA implemented in Matlab R2006a.

#### 4.2.1 Preprocessing phase

This phase consists to improve the quality of the image using filters. For this purpose, the agent UA proposes three operators. These operators are predefined in Matlab by: **'medfilt2'; 'ordfilt2'; 'wiener2 '.**
**'medfilt2'** is a 2D nonlinear operator called median filtering. It is often used in image processing to reduce "salt and pepper" noise.
**'ordfilt2'** is also a 2D nonlinear operator but more general than **'medfilt2'**. It is a 2D order-statistic filtering. B = ordfilt2 (A,order,domain) It replaces each element in the image by the orderth element in the sorted set of neighbors specified by the nonzero elements in domain.
**'wiener2 '** is a 2D adaptive noise-removal filtering. wiener2 lowpass-filters a grayscale image that has been degraded by constant power additive noise. wiener2 uses a pixelwise adaptive Wiener method based on statistics estimated from a local neighborhood of each pixel.
The agent UA proposes just one parameter to adjust for all these operators: the size of the used filter.
We define an operator by its name, number of parameters and the list of their possible values.

**Op= {operator name, numbre of parameters, List of possible Values}**

The operators of the preprocessing phase are then defined as:
```
Op = {{'medfilt2'} {1} {[3 5]}};
Op = {{'wiener2'} {1} {[3 5]}};
Op = {{'ordfilt2'} {1} {[3 5]}};
```
The alone parameter that we use is the size of the filter $t_w$. It can take two possible values: $t_w=3$ or $t_w=5$.

#### 4.2.2 Processing phase

This phase consists of detecting edges in the image. The agent UA proposes **'edge'**, a predefined operator in Matlab, as one operator for this phase.
**'edge'** takes an grayscale image I as its input, and returns a binary image BW of the same size as I, with 1's where the function finds edges in I and 0's elsewhere. For this operator, the agent UA proposes two parameters o adjust: the filter to select (sobel, prewitt, zerocross, log) and the threshold to remove edges with poor contrast [0.02,..., 0.09, 0.1]. Contours are formed by pixels higher than a given threshold.
The operator **'edge'** is then defined as:

```
Op= {{'edge'} {2} {{'sobel' 'prewitt'
'zerocross' 'log'} ; [0.02 0.03 0.04
0.05 0.06 0.07 0.08 0.09 0.1]}};
```

#### 4.2.3 Post processing phase

This phase consists of refining the image by deleting small objects. The agent UA proposes **'bwareaopen'**, a predefined operator in Matlab, as one operator for this phase.
**'bwareaopen'** is a morphological operator which remove small objects; having a connectivity inferior than a predefined threshold. It has two parameters to adjust: the connectivity and the maximal size of the objects to remove. The connectivity is defined by the number of neighbors to consider 4 or 8. For 2D images, the default connectivity is 8.
**'bwareaopen'** is then defined as:
```
Op = {{'bwareaopen'} {2} {[5 10 15
20]; [8]}};
```

### 4.3 The agent OA

After receiving all the necessary information from the agent UA, the agent OA constructs all possible combinations of operators. As the agent UA determines three phases to accomplish the segmentation task, each one of these combinations will contain three operators.
The constructed combinations are then:
C1= (medfilt2, edge, bwareaopen)
C2= (wiener2, edge, bwareaopen)
C3= (ordfilt2, edge, bwareaopen)
Each one of these operators has some parameters to adjust. For each combination, the agent OA generates an agent PA to adjust these parameters according to the dataset of images proposed by the agent UA.
There are then three agents PA1, PA2 and PA3 which treat respectively the combinations: C1, C2 and C3.

### 4.4 The agent PA

The agent PA is, generally, charged to adjust parameters of each operator in order the segmentation result will be as close as possible to the segmentation done manually by an expert.
To adjust parameters of each operator, the agent PA uses reinforcement learning. It must then define actions, states and reward. See the section 3.3.

*Actions:* are all possible combinations of parameters values. We select another action by choosing other parameters values.
An example of an action of the agent PA1: Action= [3, ('sobel', 0.02), (5,8)]

*States:* are defined by some features extracted from the image. In this application, we define a state by three features:

$$s = [\chi_1, \chi_2, \chi_3]$$

$\chi_1$ is the ratio between the number of contours in the resulting image and the number of those in the reference image (ground truth);

$\chi_2$ is the ratio between the total of white pixels in the resulting image and those in the reference image.

$\chi_3$ is the ratio between the length of the longest contour and the length of the longest contour in the reference image.

*Reward:* is defined by a weighted sum of three error measures which give some global indices about the quality of boundary-based segmentation: over detection error, under-detection error and localization error [30]. These criteria evaluate a result of edge detection. The weights used in the definition of the reward are chosen according to the importance of each criterion. The reward is defined as:

$$D = w_1 D_1 + w_2 D_2 + w_3 D_3$$

where $w_i$ is a weight for $D_i$
and

$$D_1 = ERR_{sur}(I_r, I_{ref}) = \frac{card(I_r) - card(I_r \cap I_{ref})}{card(I_r) - card(I_{ref})}$$

and

$$D_2 = ERR_{sous}(I_r, I_{ref}) = \frac{card(I_{ref/r}^{cont})}{card(I_{ref}^{cont})}$$

and

$$D_3 = ERR_{loc}(I_r, I_{ref}) = \frac{card(I_{ref/r} \cup I_{r/ref})}{card(I_r)}$$

More details about $D_1$, $D_2$ and $D_3$ are given in [30].

For each combination of operators, the agent PA finds the best parameters values which give the best segmentation. Our goal is to propose a processing method which adapts automatically to all type of images and to any vision task. In other words, for example to segment areal images the proposed approach gives, among several, the best appropriate combination of operators. Our method is aimed to be theoretically general as possible.

In this experience, we test the proposed architecture to segment two different types of images. The first one is a dataset of 70 images of traffic signs and the second contains 60 real and highly textured images. Each image has its ground-truth. Fig.5 shows the result of segmentation for 5 images taken randomly from the processing of the dataset.

It is important to note that we evaluate an operator on the whole of the dataset of images and not image by image. The fixation of some parameters of the Q- learning algorithm affects largely the result. The results showed in fig. 5 are for: α=0.5, γ=0.8, ε=0.5, nbr_épisode=200 and nbr_étapes=80.

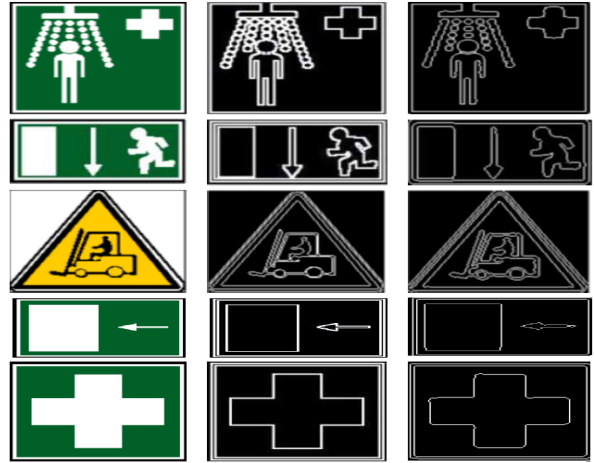

Fig. 5: from left to right: the initial image, the image segmented manually and the result of the proposed approach.

The combination of operators having the highest quality of segmentation is (wiener2,edge, bwareaopen) and the most rewarded action is (5;(prewitt,3.000000e-002);(10,8)). It is the combination of operators decided by the agent OA and returned to the agent UA. Faced to any image from the same family (traffic signs), the user can execute directly this combination of operators with their parameters values. Making use of this result we segmented 30 images of traffic signs. Results are very satisfactory. Fig. 6 shows the results of segmentation for three images from the same class of images (traffic signs) taken randomly from the processing.

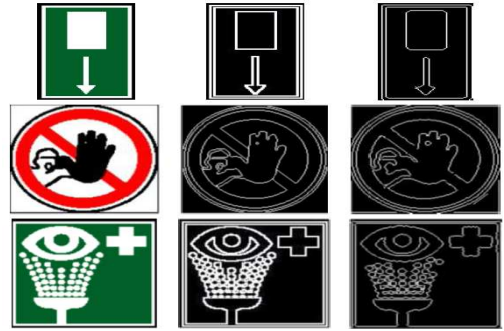

Fig. 6: results of segmentation using the combination of operators founded by the agent OA

Thus, our system finds among the proposed operators, the best ones with their optimal parameters values to apply and in which order. If the agent UA changes some information, like the set of the proposed operators for each phase, the final result changes also.

To segment images of traffic signs, our architecture finds the best combination of operators to apply. Using this combination to segment textured images we obtain bad results fig 7. The proposed operators for each phase are poor. Thus, for each type of images there are specific operators for segmentation. Then, to find a "good" segmentation for real textured images, the agent UA must feed the phases of processing by other operators and extend the list of parameters values. These operators consist of inundating textured zones.

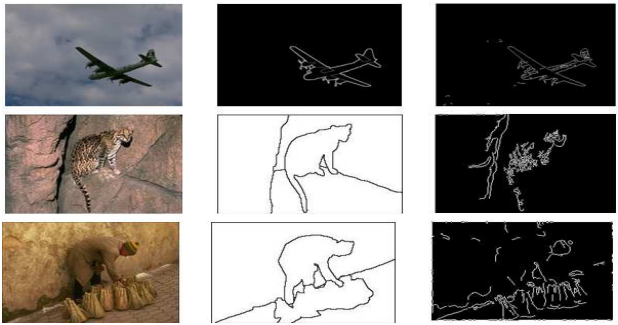

Fig. 7: from left to right: original image, manual segmentation (ground truth) and the result of the proposed approach.

The examples we have treated here validate practically our approach for simple images. For a real application, we must study the type of segmentation (boundary-based, region-based or cooperation). We must also define the appropriate operators for each phase and the right parameters values. Our approach constitutes a new general way of reasoning for any vision task that requires the right choice of operators and the right adjustment of their parameters.

Below an enlarged list of the proposed operators, as they are predefined in Matlab, and fig. 8 improvements that they bring on the textured images shown in fig. 7.

The predefined operators in Matlab and parameters values used in learning:
**medfilt2;** one parameter with two values: (3,5)
**wiener2;** one parameter with two values: (3,5)
**ordfilt2;** one parameter with two values: (3,5)
**edge;** two parameters with values: (canny, log, sobel, prewitt) and ( 0.3,0.4,0.5,0.6,0.7)
**imdilate;** two parameters with values: (line) and (3,5)
**imfill;** one parameter with one value : (holes)
**imerode;** two parameters with values : (diamond) and (1)
**bwperim;** one parameter with one value : (1)
**bwareaopen;** two parameters with values: (5,10,15):(8)

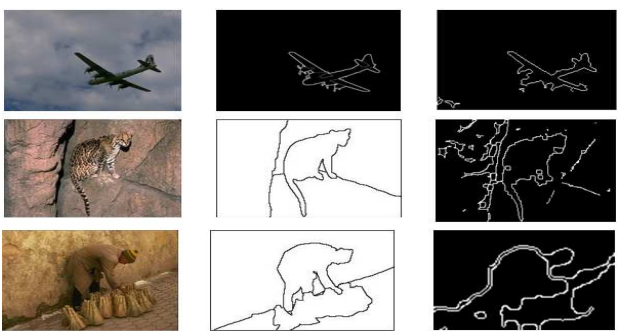

Fig. 8: Improvements after changing the method of segmentation and enlarging the list of the proposed parameters

## 5. Conclusion

Choosing the appropriate operators to apply and then adjusting their parameters values to accomplish a vision task represent a big challenge for users. In this paper we presented a multi-agents architecture based on reinforcement learning, which helps users by proposing them the optimal series of operators to apply and their best parameters values. Our system proceeds automatically to decide for his choices. Through the reinforcement learning mechanism, our architecture dose not considers only the system opportunities but also the user preferences. We intended to propose a general new way of thinking about the automatic selection of operators and the automatic adjustment of their parameters without the user intervention. The proposed approach constitutes then a theoretical robust basis for vision users and not a solution for a particular problem. The experience we have done does not restrict the application of the approach to the image processing field, but its theoretical procedure shows that it can be applied to any decision process using parametric methods. Despite the theoretical strength of the idea and the obtained results, we acknowledge that we must improve the learning algorithm and study the reward expression using a function based on the similarity between the resulting images and the ground-truth.